\def\year{2022}\relax
\documentclass[letterpaper]{article} 
\usepackage{aaai22}  
\usepackage{times}  
\usepackage{helvet}  
\usepackage{courier}  
\usepackage[hyphens]{url}  
\usepackage{graphicx} 
\usepackage{natbib}  
\usepackage{caption} 
\DeclareCaptionStyle{ruled}{labelfont=normalfont,labelsep=colon,strut=off} 
\usepackage{algorithm}

\usepackage{multirow}
\usepackage{amsmath}
\usepackage{booktabs}

\usepackage{algorithmicx}
\usepackage{algpseudocode}
\usepackage{newfloat}
\usepackage{listings}
\floatstyle{ruled}
\newfloat{listing}{tb}{lst}{}
\floatname{listing}{Listing}
\title{Reducing Flipping Errors in Deep Neural Networks}

\title{Reducing Flipping Errors in Deep Neural Networks \footnote{Code: https://github.com/Xiang-Deng-DL/FER} }

\author {
    Xiang Deng\textsuperscript{\rm 1}\footnote{Interns at JD.com.},
    Yun Xiao\textsuperscript{\rm 2},
    Bo Long\textsuperscript{\rm 2},
    Zhongfei Zhang\textsuperscript{\rm 1}
}
\affiliations {
    \textsuperscript{\rm 1} Computer Science Department, State University of New York at Binghamton\\
    \textsuperscript{\rm 2} JD.com\\
    xdeng7@binghamton.edu, xiaoyun1@jd.com, bo.long@jd.com, zhongfei@cs.binghamton.edu
}

\usepackage{bibentry}

\begin{document}

\maketitle

\begin{abstract}
Deep neural networks (DNNs) have been widely applied in various domains in artificial intelligence including computer vision and natural language processing.
A DNN is typically trained for many epochs and then a validation dataset is used to select the DNN in an epoch (we simply call this epoch ``the last epoch") as the final model for making predictions on unseen samples, while it usually cannot achieve a perfect accuracy on unseen samples.
An interesting question is ``how many test (unseen) samples that a DNN misclassifies in the last epoch were ever correctly classified by the DNN before the last epoch?".
In this paper, we empirically study this question and find on several benchmark datasets that the vast majority of the misclassified samples in the last epoch were ever classified correctly before the last epoch, which means that the predictions for these samples were flipped from ``correct" to ``wrong".
Motivated by this observation, we propose to restrict the behavior changes of a DNN on the correctly-classified samples so that the correct local boundaries can be maintained and the flipping error on unseen samples can be largely reduced.
Extensive experiments on different benchmark datasets with different modern network architectures demonstrate that the proposed flipping error reduction (FER) approach can substantially improve the generalization, the robustness, and the transferability of DNNs without introducing any additional network parameters or inference cost, only with a negligible training overhead.
\end{abstract}

\section{Introduction}
The superior performances of deep neural networks (DNNs) have driven their wide deployment in varieties of applications in artificial intelligence such as medical diagnosis \cite{miotto2016deep}, automated vehicle control \cite{levinson2011towards}, and financial business \cite{ozbayoglu2020deep}.
Despite the remarkable performances, it is appealing and necessary to further improve them, thus benefiting these applications and extending the application horizon to more accuracy-critical or safety-critical domains.
However, furthering improving the performance of a DNN usually needs to substantially, even exponentially, increase the number of its parameters \cite{simonyan2014very, he2016deep}, which leads to a large amount of computation and memory costs.
In this work, we aim to enhance the performance of a DNN classifier without introducing any additional network parameters, without increasing inference cost, and almost without training overhead by exploring the learning dynamics of DNNs.\par

\begin{figure}[t]
\centering
\includegraphics[width=0.47\textwidth]{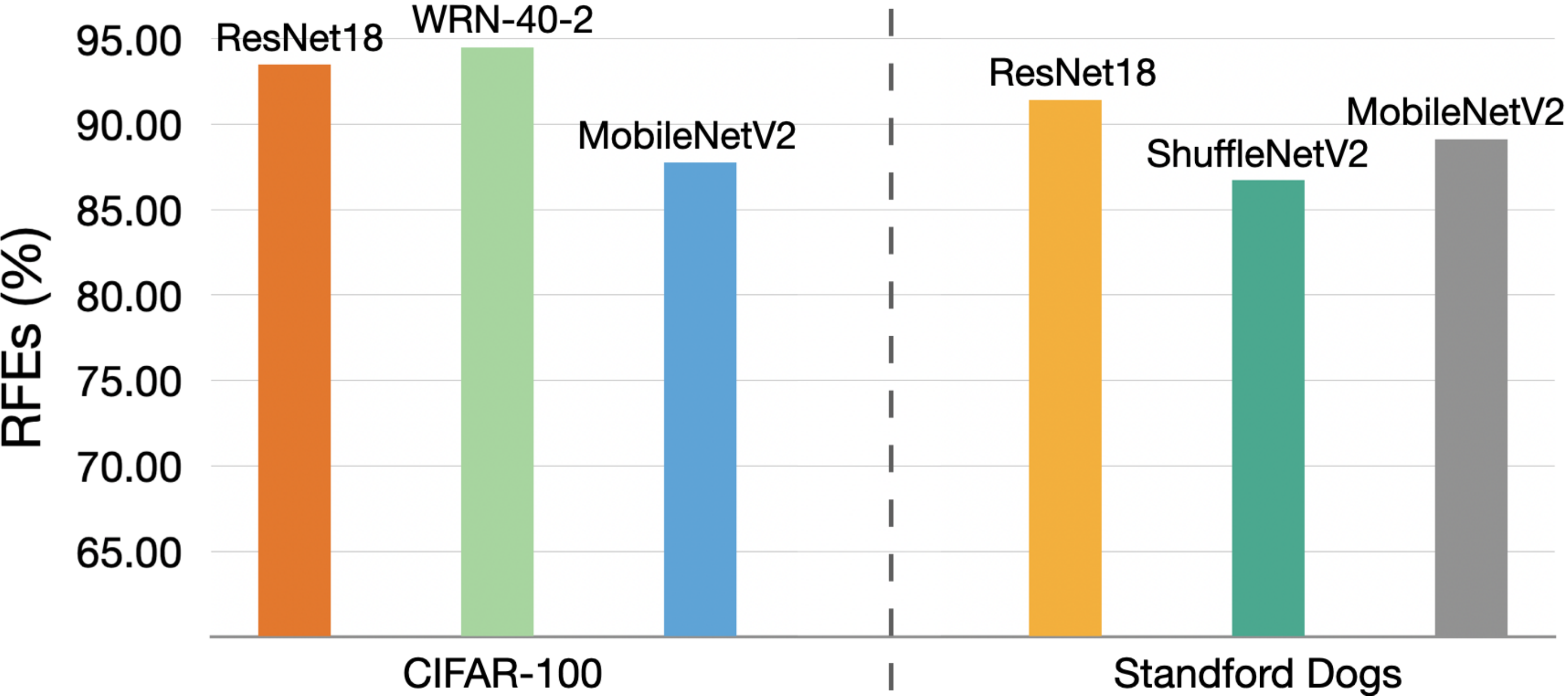}
\caption{RFEs of different DNNs on CIFAR-100 and Standford Dogs.}
\label{rfe}
\vspace{-0.2in}
\end{figure}

Currently, DNNs are typically trained for many epochs and then a validation dataset is used to select the DNN in the best epoch as the final model for making predictions on unseen samples or the validation dataset is used to set up the total number of training epochs and then the DNN at the last epoch is selected as the final model.
Without losing generality, we simply denote the selected epoch in both cases by the last epoch.
An interesting question regarding the generalization of a DNN classifier is ``how many misclassified test samples in the last epoch were ever correctly classified by the DNN before the last epoch?".
To quantitatively study this question, we first introduce several novel metrics and definitions:
(1) \textbf{Wrongly Flipped Samples (WFSs)} are the samples that are misclassified by the DNN in the last epoch but were ever classified correctly by the DNN before the last epoch;
(2) \textbf{Relative Flipping Error (RFE)} is the ratio of WFSs to the total number of misclassified test samples (MTSs) in the last epoch, i.e., $RFE = \frac{\#WFSs}{\#MTSs}$.
We report the RFEs of various modern networks on two different benchmark datasets including a regular classification dataset, i.e., CIFAR-100 \cite{krizhevsky2009learning} and a fine-grained classification dataset, i.e., Standford Dogs \cite{khosla2011novel}. As shown in Figure \ref{rfe}, across different modern network architectures on both datsets, a large percentage of the wrongly-classified test samples in the last epoch were ever correctly classified before the last epoch, e.g., the RFEs of ResNet-18 are over 90\% on both datasets. 
This indicates that the DNN has ever learned the correct local boundaries for these samples but then flipped them to the wrong directions.
\par

Inspired by the observation, we propose a novel framework, i.e., flipping error reduction (FER), to reduce the number of wrongly-flipped samples and thus to enhance DNN performances.
FER is based on the idea that once a DNN has made a correct prediction on a sample, we limit the behavior change of the DNN on this sample with the goal of maintaining the correct local classification boundaries near this sample to avoid wrong flipping.
To capture the local boundaries, we define the behavior of a DNN on a sample as its output soft distribution over different classes which captures the distances between a sample and different class boundaries.
Since a DNN may make correct predictions in many epochs for a sample, we maintain a weighted average of these correct behaviors based on their confidences and epoch numbers.
A behavior with a higher confidence or in a later epoch contributes more to the average.
The weighted average behavior is more stable and accurate than a single behavior for restricting the DNN behavior changes on well-learned samples.

FER is a simple and easy-to-implement yet effective approach for training DNNs.
Extensive experiments on different types of benchmark datasets demonstrate that FER can improve the generalization, the robustness, and the transferability of DNNs without introducing any additional network parameters, without increasing inference cost but only with a negligible training overhead for maintaining the average behaviors of a DNN on correctly-classified samples.\par

Our contributions are summarized as follows:
\begin{itemize}
\item We study a novel problem about "how many wrongly-classified test samples in the last epoch were ever correctly classified before the last epoch?" and define novel metrics to quantitatively investigate this question and further propose a solution to this problem.
To our best knowledge, this is the first work to address this issue.

\item 
To reduce flipping errors, we propose a novel framework, i.e., FER, that restricts the behavior changes of a DNN on correctly-classified samples and thus maintains the correct local boundaries near these samples.

\item We empirically show on different types of benchmark datasets that FER is able to improve the generalization, the robustness, and the transferability of modern DNNs, without introducing any network parameters or inference cost, only with a negligible training overhead.
\end{itemize}

\section{Related Work}
In this paper, we study the wrongly flipped predictions on unseen samples.
This phenomenon also happens to training (seen) samples \cite{toneva2018empirical}.
However, as current DNNs are powerful enough to memorize the whole training dataset, these wrongly-flipped predictions on training samples can almost always be corrected by the DNNs through memorization in the later epochs.
Memorization does not mean generalization and thus the wrong flipping issue on unseen samples is more challenging.
FER is designed to solve this problem by restricting the soft output distribution changes of a DNN on correctly-classified samples to maintain local correct classification boundaries.
Hence, it is technically related to label smoothing \cite{szegedy2016rethinking} and knowledge distillation (KD) techniques \cite{hinton2015distilling,deng2021comprehensive,deng2021graph} which are also based on soft distribution.

\textbf{Label Smoothing} Szegedy et al. \shortcite{szegedy2016rethinking} propose the label smoothing regularizer (LSR) that linearly interpolates between a one-hot label and a uniform distribution to generate a soft distribution as the label for training DNNs, and empirically show that it is able to improve the performance of the Inception architecture on image classification.
Ever since then, LSR has been used in various advanced image classification models \cite{zoph2018learning,real2019regularized,huang2019gpipe} and has also been introduced to other domains.
In speech recognition, \citeauthor{chorowski2016towards} \shortcite{chorowski2016towards} show that label smoothing reduces the word error rate on the WSJ dataset.
In machine translation, Vaswani et al. \shortcite{vaswani2017attention} show that label smoothing improves the BLEU score but hurts the perplexity.
Muller et al. \shortcite{muller2019does} empirically demonstrate that LSR leads to a better calibration for DNNs.
Recently, Yuan et al. \shortcite{yuan2020revisiting} propose a new manually designed label smoothing based regularizer, i.e., TF-Reg, that combines LSR and a temperature.
These label smoothing techniques manually assign the same small probability to the non-ground-truth classes and thus fail to take into account sample-to-class similarities and cannot fully utilize the advantages of soft labels.

\begin{figure*}[t]
\centering
     \includegraphics[width=0.95\textwidth]{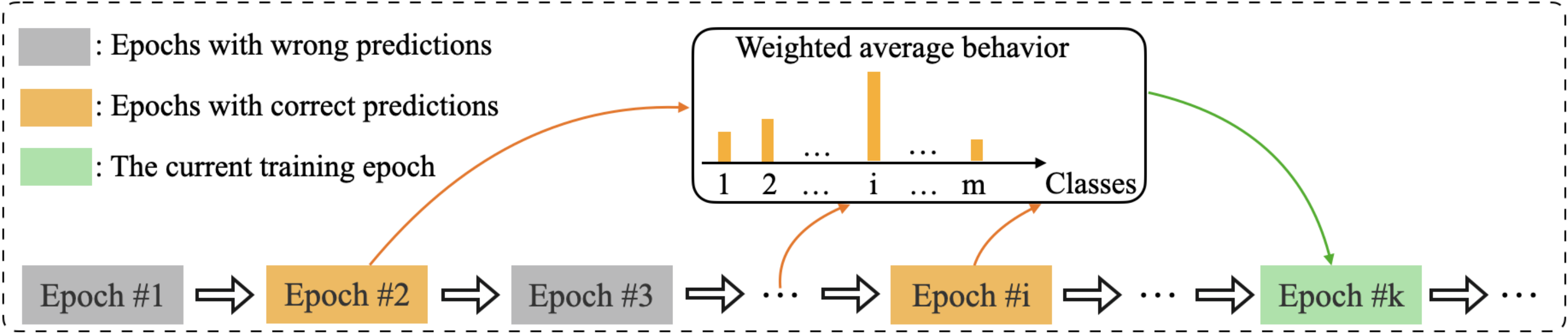}
     \caption{Framework of FER.}
     \label{framework}
     \vspace{-0.1in}
\end{figure*}

\textbf{Knowledge Distillation} Knowledge distillation (KD) \cite{hinton2015distilling} overcomes the limitations of the manually designed soft labels by taking advantage of a pretrained teacher network to generate soft labels which contain sample-to-class similarity (distance) information.
However, it also brings several times of training cost over the manually designed soft labels due to the two facts: (1) KD first requires training a large DNN as the teacher; (2) when training the student, KD needs to process each sample twice in each iteration, once by the teacher and once by the student.
To reduce the cost of training a large teacher, many self-distillation approaches including but not limited to \cite{xu2019data,zhang2019your,yang2019snapshot,yang2019training,furlanello2018born,bagherinezhad2018label,yun2020regularizing,deng2021learning,ji2021refine} have been proposed.
Zhang et al. \shortcite{zhang2019your} and Ji et al. \shortcite{ji2021refine} add additional layers or parameters to a DNN to generate soft labels, which improves the performance but introduces large computation and memory cost.
Born-again networks \cite{furlanello2018born,yang2019training} and label-refine networks \cite{bagherinezhad2018label} train a DNN for many generations and the network in the $(i-1)$th generation is used as the teacher to train the network in the $i$th generation.
In other words, these approaches avoid pretraining a large teacher network by pretraining the network itself as its own teacher, and this process can be repeated many times.
Nevertheless, they still have a large training overhead as they need to train a network for many times.
To further reduce the training cost for multiple-generation training, SD \cite{yang2019snapshot} introduces mini-generation training.
It borrows the idea from \cite{huang2017snapshot} to rely on a cyclic learning rate schedule \cite{loshchilov2016sgdr} to train DNNs for many mini-generations.
However, this learning rate schedule causes that SD cannot update the soft labels frequently and introduces wrong supervision signals from mini-generations.
To free the update frequency of soft labels, LWR \cite{deng2021learning} uses the sample-to-class similarities learned in a past epoch in a standard training procedure to generate soft labels.
However, it still introduces wrong supervision signals for misclassified samples and the soft labels generated from the information in one epoch are unstable and less accurate.
FER addresses these issues by limiting the behavior changes of a DNN on correctly-classified samples and taking into account the DNN behaviors in many epochs.
\par

Note that we aim to improve the DNN performance with comparable training and inference costs to the standard training procedure through exploring the learning dynamics of DNNs.
Thus, we do not compare the proposed method with those approaches (e.g., born-again networks or KD) whose training or inference cost is several times of ours.


\section{Framework}
In this section, we introduce FER for improving DNN performances by exploring the learning dynamics of DNNs.
To show the connection between FER and the standard training procedure, we first review the standard procedure and then make further derivations to present FER.

\subsection{Standard Training Procedure}
We denote the training dataset by $\mathcal{D} = \{ (x_i, y_i) \}_{i=1}^n$ of $n$ samples where $x_i$ denotes an input sample and $y_i$ represents the corresponding one-hot label. The standard training (STD) procedure trains a DNN $f$ by minimizing the cross-entropy loss between the DNN output and the one-hot label:
\begin{equation}
\label{1}
\mathcal{L}_{CE}(x) = \mathcal{H}(f(x), y)
\end{equation}
where $\mathcal{H}$ denotes cross-entropy.

At the beginning of the training process, a DNN with initial weights makes wrong predictions on most samples.
The DNN is then trained for multiple epochs with the goal of making correct predictions on data samples.
However, we find that the behaviors of a DNN on many test samples are flipped from correct predictions to wrong predictions as the training progresses, which is opposite to the learning goal.
Inspired by this observation, we propose to improve DNN performances by reducing this kind of flipping errors.

\subsection{Flipping Error Reduction}
To reduce flipping errors, we propose to regularize the behavior changes of a DNN on the samples that are already classified correctly.
Note that we do not restrict the behavior of a DNN on the misclassified samples, since the local boundaries near these samples are still wrong and cannot be used to make correct predictions on unseen samples.
The behavior of a DNN on a sample is defined as a soft distribution:
\begin{equation}
\label{b2}
b(x) = \sigma( \frac{f(x)}{\tau} )
\end{equation}
where $\sigma$ denotes the softmax function and $\tau$ is a temperature to soften the outputs.
The advantage of using a soft distribution is that soft labels are able to capture the sample-to-class-margin distances, i.e., sample-to-class similarities.
By restricting the behavior changes of a DNN on correctly-classified samples, the correct local boundaries near these samples can be maintained, thus reducing wrong flipping.\par

Since a DNN is typically trained for many epochs, there can be multiple correct behaviors of a DNN on a sample.
These correct behaviors are generated in different epochs and may also have different confidences.
Here the confidence means the probability for the ground-truth class, i.e., $c= \sigma(f(x))[\textbf{y}]$ where $\textbf{y}$ is the ground-truth class index.
A natural question is ``which correct behavior should be used for restricting the DNN behavior on a sample?".
As shown in  Figure \ref{framework}, we maintain a weighted average of all these correct behaviors based on the confidence and the epoch number of each behavior for each sample.
Specifically, in epoch $k$, suppose that the DNN has made correct predictions in $j$ epochs for sample $x$ where $0\leq j\leq k-1$.
We denote the behaviors in these $j$ epochs with correct predictions by $[b_1(x), b_2(x), ..., b_j(x)]$ where $b_i(x)$ is the behavior of the DNN on sample $x$ in the $i$th behavior-correct epoch.
For the behavior in a later epoch or with a high confidence, we assign a larger weight as the DNN has learned more information or is more confident.
The weighted average behavior in epoch $k$ is calculated with:
\begin{equation}
\label{3}
\mathbf{\hat{b}}_k(x) = \mathrm{e}^{-\mu c_j}\mathbf{\hat{b}}_{k-1}(x) + (1-\mathrm{e}^{-\mu c_j})b_j(x)
\end{equation}
where $c_j$ is the confidence of $b_j(x)$ and $\mu$ is a coefficient to scale the contribution of confidence $c_j$.
Besides $b_j(x)$, all the other behaviors with correct predictions also contribute to the current average behavior $\mathbf{\hat{b}}_k(x)$ as they are involved in computing $\mathbf{\hat{b}}_{k-1}(x)$.
We then restrict the behavior changes of the DNN on correctly-classified samples with a regularizer:
\begin{equation}
\label{eq4}
\mathcal{L}_{cor}(x) = \alpha \mathcal{L}_{CE}(x) + \beta \mathcal{K}(\sigma( \frac{f(x)}{\tau} ), \mathbf{\hat{b}}_k(x))
\end{equation}
where $\alpha$ and $\beta$ are two balancing weights; $\mathcal{K}$ denotes KL-divergence.
The gradients do not flow through $\mathbf{\hat{b}}_k(x)$ as it is the record of the past behaviors of a DNN.
In most cases, we simply set $\alpha$ and $\beta$ to $1.0-0.9\frac{k}{M}$ and $0.9\frac{k}{M}$ (where $M$ is the number of total training epochs), respectively, based on the fact that the average behavior becomes more accurate in the later epoch with the DNN learning more information.\par

FER restricts the behaviors of a DNN on correctly-classified samples but gives the totally free space for the DNN on learning misclassified samples, thus the final objective is written as:
\begin{equation}
\label{5}
\mathcal{L}_{FER}(x) = \mathbf1_{[T(x)]} \mathcal{L}_{cor}(x)+(1-\mathbf1_{[T(x)]})\mathcal{L}_{CE}(x)
\end{equation}
where $\mathbf1_{[.]}$ is the indicator function which equals to one if the argument in the subscript [.] is true and zero otherwise; $T(x)$ denotes the argument that sample $x$ has been ever classified correctly before the current epoch.
As we can see, the training overhead of FER is the cost for maintaining a soft behavior average (i.e., $\mathbf{\hat{b}}_k(x)$) for each correctly-classified sample, which is negligible as these soft behaviors do not need to be stored in the GPU memory.

\subsection{Analysis of FER}
\begin{figure}[t]
\centering
\includegraphics[width=0.47\textwidth]{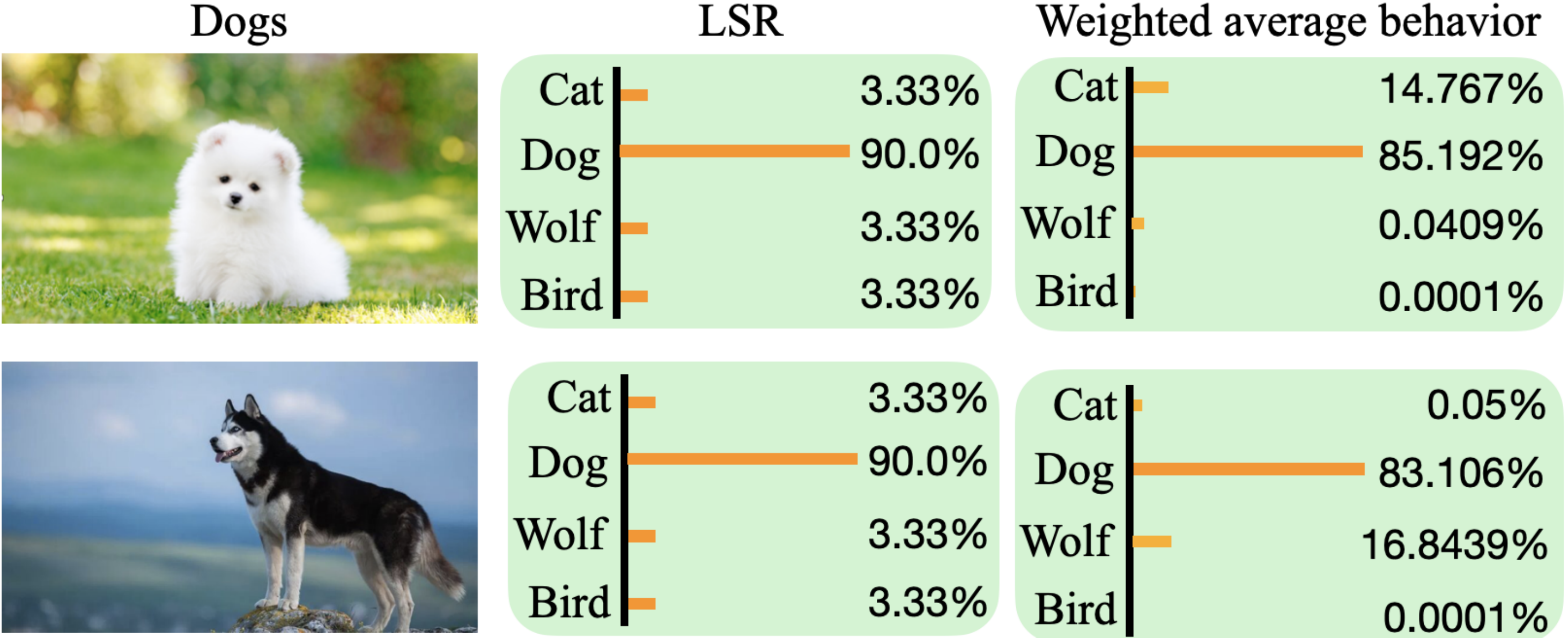}
\caption{Due to the body type differences, the dog in the first row is more visually similar to a cat while the the dog in the second row is more visually similar to a wolf.
Different from LSR that assigns them the same soft label, FER uses the weighted average behavior that assigns different probabilities to different classes.}
\label{soft}
\end{figure}

\paragraph{Label Smoothing Effects} FER is highly related to LSR as it defines the DNN behavior as a soft distribution.
LSR assigns a small probability to the non-ground-truth classes by using the weighted average of the one-hot label $y$ and a uniform distribution as the training targets, i.e., $y_{LSR}= (1-\epsilon) y + \frac{\epsilon}{m}$ where $\epsilon$ is a manually set hyper-parameter and usually set to 0.1; $m$ is the total number of classes.
The objective of LSR is thus expressed as:
\begin{equation}
\label{6}
\begin{split}
\mathcal L_{LSR} = \mathcal{K}(\sigma (f(x)), y_{LSR})        \\=(1-\epsilon)\mathcal{L}_{CE}(x)+\epsilon \mathcal{K}(\sigma(f(x), \frac{\epsilon}{m}))
\end{split}
\end{equation}
It is observed that (\ref{6}) shares a similar form to (\ref{eq4}) that is the loss for the correctly-classified samples in FER.
FER is thus reduced to LSR when $\mathbf{\hat b}(x)$ follows a manually-set uniform distribution and the regularizer is put on all samples by ignoring whether the predictions on them are already correct or not.
This indicates that FER is an adaptive version of LSR and should inherit the advantage of LSR such as a better generalization.
The advantage of FER over LSR is that FER can capture the sample-to-class distance information which is accumulated from the past behaviors of the DNN, while LSR fails to provide this kind of information as it assigns the same small probability to all the non-ground-truth classes.
To further illustrate this point, we provide an intuitive example in Figure \ref{soft}.

\paragraph{Maintaining Correct Local Classification Boundaries} FER defines the the DNN behavior as the soft output distribution over different classes which contains sample-to-class distance information.
It restricts the DNN behavior changes on the correctly-classified samples and gives free space for learning wrongly-classified samples.
FER thus mitigates wrong flipping through maintaining the correct local boundaries while allowing correct flipping (i.e., from wrong predictions to correct predictions) for wrongly-classified samples.
We empirically validate the flipping error reduction effects of FER in the ablation studies.


\section{Experiments}
We first conduct ablation studies and then compare FER with cost-comparable label-smoothing approaches.

\begin{figure*}[!t]
       \begin{minipage}{0.48\textwidth}
      \centering
     \includegraphics[width=0.85\textwidth]{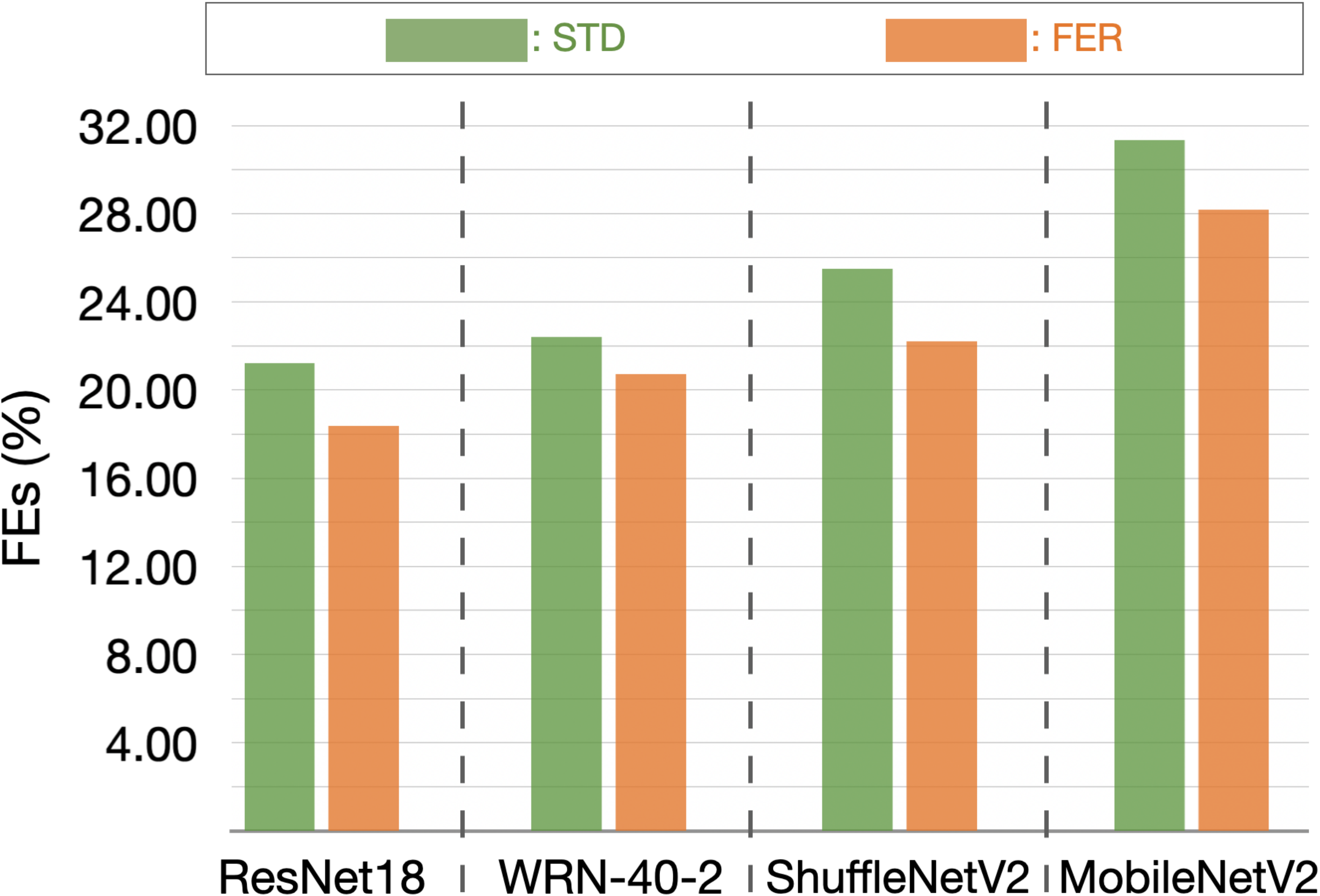}     \caption{Flipping errors on CIFAR-100.}
     \label{fe_cifar100}
   \end{minipage}\hfill
   \begin{minipage}{0.48\textwidth}
\centering
     \includegraphics[width=0.85\textwidth]{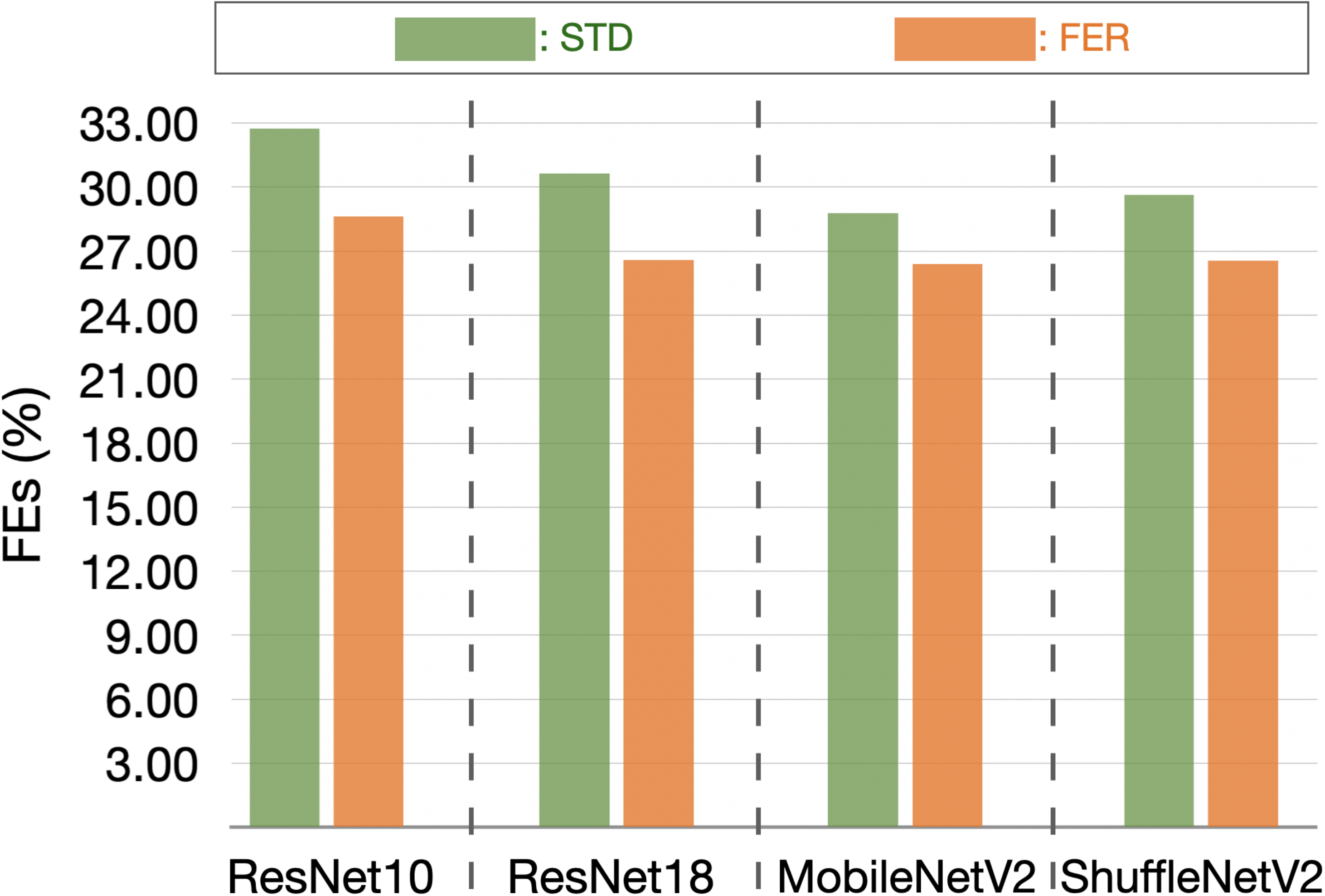}
     \caption{Flipping errors on Standford Dogs.}
     \label{fe_dogs}
   \end{minipage}\hfill
\end{figure*}

\begin{table}[!t]
\centering
\begin{tabular}{llll}
\toprule
                     & \multicolumn{1}{l}{} &ShuffleNetV2    &ResNet-18     \\\midrule
\multirow{4}{*}{\begin{tabular}[c]{@{}l@{}}CIFAR \\ -100\end{tabular}} 
                     &FER    &\textbf{74.82$\pm$0.24}   &\textbf{80.29$\pm$0.12} \\
                     &FER w/o S1    &74.48$\pm$0.33 &80.00$\pm$0.18 \\
                     &FER w/o S2    &74.46$\pm$0.19  &79.90$\pm$0.07  \\
                     &LWR  &73.53$\pm$0.33 &79.73$\pm$0.32  \\
                     \midrule \midrule

\multirow{4}{*}{\begin{tabular}[c]{@{}l@{}}Standford \\Dogs\end{tabular}} 
                    &FER &\textbf{68.76$\pm$0.17}    &\textbf{70.64$\pm$0.25}       \\
                &FER w/o S1    &68.01$\pm$0.40 &70.25$\pm$0.25\\
                &FER w/o S2  &67.82$\pm$0.10 &70.08$\pm$0.18 \\
                &LWR  &67.39$\pm$0.42  &69.84$\pm$0.45\\ \bottomrule
\end{tabular}
\vspace{-0.05in}
\caption{Ablation studies regarding the strategies in FER.}
\label{strategy}
\end{table}

\begin{table}[!t]
\centering
\begin{tabular}{cccc}
\toprule
      &$\mu$ = 0.5           &$\mu$ = 1.0       &$\mu$ = 1.5       \\\midrule
$\tau$ = 5 & 79.95$\pm$0.14 &80.29$\pm$0.12 &79.91$\pm$0.16\\
$\tau$ = 10 &79.40$\pm$0.33 &79.63$\pm$0.15 &79.64$\pm$0.09
\\ \bottomrule  
\end{tabular}
\vspace{-0.05in}
\caption{Effects of $\tau$ and $\mu$ with ResNet-18 on CIFAR-100.}
\label{hyper}
\vspace{-0.2in}
\end{table}

\begin{table*}[!t]
\centering
\resizebox{\textwidth}{!}{%
\begin{tabular}{llllll}
\toprule
                     & \multicolumn{1}{l}{} &ResNet-18    & WRN-40-2     & ShuffleNetV2   & MobileNetV2     \\\midrule
\multirow{7}{*}{\begin{tabular}[c]{@{}l@{}}CIFAR-100\end{tabular}} &STD &77.31$\pm$0.33 &76.24$\pm$0.15   & 71.58$\pm$0.19 & 64.27$\pm$0.81   \\
                     &LSR    &78.24$\pm$0.16    & 76.29$\pm$0.25 &71.90$\pm$0.13  & 64.54$\pm$0.74 \\
                      &Max-Entropy &77.65$\pm$0.35 &76.34$\pm$0.15 &71.23$\pm$0.22&64.34$\pm$0.46 \\
                     &SD     &78.31$\pm$0.24    & 76.34$\pm$0.12 &68.70$\pm$0.51 & 65.78$\pm$0.13 \\
                     &CS-KD    &78.01$\pm$0.13    &76.25$\pm$0.10  &71.43$\pm$0.28  
                     &64.81$\pm$0.20 \\
                     &TF-Reg  &77.36$\pm$0.23 &76.32$\pm$0.19  &72.09$\pm$0.34 &64.87$\pm$0.27   \\
                     &FER (Ours) &\textbf{80.29$\pm$0.12 ($\uparrow$ 2.98)} &\textbf{77.80$\pm$0.13 ($\uparrow$ 1.56)} &\textbf{74.82$\pm$0.24 ($\uparrow$ 3.24)} & \textbf{66.84$\pm$0.79 ($\uparrow$ 2.57)}\\ \midrule \midrule

\multirow{7}{*}{\begin{tabular}[c]{@{}l@{}}Tiny ImageNet\end{tabular}}  &STD   &65.57$\pm$0.16    &58.71$\pm$0.44    &60.80$\pm$0.25  &  55.53$\pm$0.76    \\
                    &LSR &64.91$\pm$0.08    &58.97$\pm$0.10    &61.85$\pm$0.27 
                    &56.04$\pm$0.41
                        \\
                   &Max-Entropy &65.25$\pm$0.16 &58.82$\pm$0.22 &61.56$\pm$0.40&55.92$\pm$0.42  \\
                &SD    &65.87$\pm$0.28    &61.01$\pm$0.10   &60.79$\pm$0.15  &55.97$\pm$0.59   \\
                &CS-KD   &64.29$\pm$0.25    &59.67$\pm$0.13  &61.66$\pm$0.34  & 57.08$\pm$0.52          \\
                &TF-Reg  &64.88$\pm$0.47    &58.62$\pm$0.49     &61.55$\pm$0.42  &56.03$\pm$0.23    \\
                &FER (Ours) & \textbf{67.72$\pm$0.32 ($\uparrow$ 2.15)} &\textbf{62.10$\pm$0.15 ($\uparrow$ 3.39)} &\textbf{62.61$\pm$0.04  ($\uparrow$ 1.81)} &\textbf{57.43$\pm$0.23 ($\uparrow$ 1.90)}  \\ \bottomrule
\end{tabular}
}
\caption{Test accuracies (\%) on CIFAR-100 and Tiny ImageNet. $\uparrow$ denotes the absolute improvement over the standard training procedure (i.e., STD).}
\label{regular}
\vspace{-0.2in}
\end{table*}
\subsection{Ablation Studies}
\subsubsection{Can FER Reduce Flipping Errors?}
FER is developed from the STD training procedure with the goal of reducing the number of wrongly flipped samples.
In this part, we aim to empirically validate that FER indeed reduces the flipping errors of DNNs.
The \textbf{flipping error (FE)} is quantitatively defined as the ratio of the number of wrongly flipped test samples (WFSs) to the total number of test samples (TSs):
\begin{equation}
\label{fe}
FE=\frac{\#WFSs}{\#TSs}
\end{equation}
We conduct experiments on two different types of datasets including one regular classification dataset and one fine-grained dataset, i.e., CIFAR-100 \cite{krizhevsky2009learning} and Standard Dogs \cite{khosla2011novel}.

Figure \ref{fe_cifar100} and Figure \ref{fe_dogs} report the FEs on CIFAR-100 and Standford Dogs, respectively.
It is observed that by simply restricting the DNN behaviors on correctly-classified samples, FER consistently and substantially reduces the FEs across different networks and datasets, which demonstrates the ability of FER to mitigate wrong flipping.

\subsubsection{FER v.s. LWR}
There are two strategies in FER: (1) only restricting the DNN behavior changes on correctly-classified samples while giving the DNN totally free space for learning on misclassified samples, instead of restricting the DNN behavior on all the samples; (2) using the weighted average of multiple correct behaviors instead of one single behavior to restrict a DNN.
In this part, we empirically examine the effects of the two strategies. 
We denote FER without strategy (1) by \textbf{FER w/o S1}, without strategy (2) by \textbf{FER w/o S2}.
Without both of the two strategies, FER is reduced to LWR \cite{deng2021learning} that is a special case of FER.
\par

Table \ref{strategy} reports the performances on CIFAR-100 and Standford Dogs.
As expected, without either of the two strategies or without both (i.e., LWR), the performances drop significantly on both datasets, which demonstrates the effectiveness of the two strategies and the superiority of FER over LWR.

\subsubsection{Effects of $\tau$ and $\mu$}
$\tau$ in FER controls the softness of the behavior while $\mu$ scales the contribution of the confidence of each behavior to the weighted average.
Their effects on the performances of FER are reported in Table \ref{hyper}.
It is observed that $\tau$ = 5 consistently outperforms $\tau$ = 10.
The reason is that as seen from Eq. (\ref{b2}), a large $\tau$ makes the behavior too flat to contain sample-to-class distance information. 
On the other hand, we observe that the performance of FER is not sensitive to $\mu$.

\begin{table*}[!t]
\centering
\resizebox{\textwidth}{!}{%
\begin{tabular}{llllll}
\toprule
                               & \multicolumn{1}{l}{} & ResNet-10      & ResNet-18      & MobileNetV2    & ShuffleNetV2   \\ \midrule
\multirow{7}{*}{CUB}           & STD                  & 58.96$\pm$0.12 & 61.09$\pm$0.49 & 67.20$\pm$0.21 & 61.67$\pm$0.85 \\
                               & LSR                  & 59.31$\pm$0.21 & 63.57$\pm$0.50 & 67.97$\pm$0.43 & 62.66$\pm$0.11 \\
                            & Max-Entropy          & 59.00$\pm$0.30 & 61.23$\pm$0.37 &66.56$\pm$0.30  &61.10$\pm$0.15  \\
                               & SD                   & 59.28$\pm$0.34 & 64.19$\pm$0.24 & 68.15$\pm$0.32 & 63.99$\pm$0.29 \\
                             
                               & CS-KD               &60.70$\pm$0.21  &64.57$\pm$0.29  &67.48$\pm$0.32  &63.32$\pm$0.25   \\
                               & TF-Reg               & 58.84$\pm$0.60 &62.04$\pm$0.28 &67.20$\pm$0.43&61.19$\pm$0.58                \\
                               & FER (Ours)  &\textbf{64.86$\pm$0.45 ($\uparrow$ 5.90)} &\textbf{67.65$\pm$0.25 ($\uparrow$ 6.56)} & \textbf{69.08$\pm$0.47 ($\uparrow$ 1.88)} & \textbf{65.60$\pm$0.26 ($\uparrow$ 3.93)} \\ \midrule \midrule 
\multirow{7}{*}{Stanford Dogs} & STD                  &63.91$\pm$0.25  &66.56$\pm$0.28  &68.05$\pm$0.26 &66.08$\pm$0.32  \\
                               & LSR                  &63.36$\pm$0.03  &67.12$\pm$0.86  &69.13$\pm$0.09 &66.90$\pm$0.34 \\
                              & Max-Entropy          &63.94$\pm$0.30  &66.42$\pm$0.50  &67.97$\pm$0.30  &66.25$\pm$0.60    \\
                               & SD                   &64.65$\pm$0.36  &68.79$\pm$0.06  &70.26$\pm$0.35  &67.30$\pm$0.26       \\
                               
                               & CS-KD               &64.91$\pm$0.26  &69.17$\pm$0.19  &68.73$\pm$0.25  &66.75$\pm$0.31   \\
                               
                               & TF-Reg               &63.72$\pm$0.44  &66.53$\pm$0.50  &68.36$\pm$0.26  &66.63$\pm$0.26   \\
                               & FER (Ours)  &\textbf{67.12$\pm$0.29 ($\uparrow$ 3.21)} & \textbf{70.64$\pm$0.25 ($\uparrow$ 4.08)} &\textbf{70.78$\pm$0.31 ($\uparrow$ 2.73)}  & \textbf{68.76$\pm$0.17 ($\uparrow$ 2.68)}      \\ \midrule \midrule  
\multirow{7}{*}{FGVC-Aircraft} & STD                  &73.89$\pm$0.25  &79.58$\pm$0.25  &83.01$\pm$0.30   & 78.00$\pm$0.45     \\
                               & LSR                  &74.52$\pm$0.18  &80.91$\pm$0.28    &83.88$\pm$0.10   &78.40$\pm$0.94    \\
                               & Max-Entropy          &73.39$\pm$0.09  &79.59$\pm$0.41   &82.81$\pm$0.45    &78.20$\pm$0.25   \\
                               & SD                   &74.98$\pm$0.38  &80.55$\pm$0.80   &83.36$\pm$0.13  &78.09$\pm$0.34    \\
                               & CS-KD               &74.95$\pm$0.40  &79.72$\pm$0.19 &80.62$\pm$0.38  &77.89$\pm$1.55   \\
                               & TF-Reg               &73.69$\pm$0.38  &80.12$\pm$0.33 &83.39$\pm$0.11  &78.50$\pm$0.31   \\
                               & FER (Ours) & \textbf{77.02$\pm$0.11 ($\uparrow$ 3.13)} &\textbf{82.38 $\pm$0.10 ($\uparrow$ 2.80)} &\textbf{84.56$\pm$0.82 ($\uparrow$ 1.55)} & 
                              \textbf{78.60$\pm$0.46 ($\uparrow$ 0.60)}       \\ \bottomrule
\end{tabular}
}
\vspace{-0.05in}
\caption{Test accuracies (\%) on fine-grained classification datasets. $\uparrow$ denotes the absolute improvement over the STD procedure.}
\label{fine}
\vspace{-0.15in}
\end{table*}

\begin{table}[!t]
\centering
\resizebox{0.47\textwidth}{!}{%
\begin{tabular}{lll}
\toprule
        &Arcene       &Iris        \\\midrule
STD   &83.00$\pm$2.16 &90.00$\pm$2.72\\
LSR   &81.00$\pm$4.55 &92.22$\pm$1.57    \\ 
Max-Ent   &79.67$\pm$3.77 &94.44$\pm$1.57 \\
SD    &81.00$\pm$1.63    &93.33$\pm$2.72 \\
CS-KD   &83.67$\pm$0.94    & 94.44$\pm$1.57\\ 
TF-Reg  &80.00$\pm$2.16    &90.00$\pm$2.72 \\
FER (Ours) &\textbf{86.33$\pm$0.94 ($\uparrow$ 3.33)} &\textbf{95.56$\pm$1.57 ($\uparrow$ 5.56)}\\ \bottomrule  
\end{tabular}
}
\vspace{-0.05in}
\caption{Test accuracies (\%) on tabular datasets.}
\label{tabular}
\vspace{-0.2in}
\end{table}

\subsection{Generalization on Test Data}
We use the test accuracy to measure the generalization of different methods on test data.

\textbf{Datasets} We adopt different kinds of datasets across different domains including two regular classification datasets, i.e., CIFAR-100 \cite{krizhevsky2009learning} and Tiny-ImageNet \footnote{https://tiny-imagenet.herokuapp.com}, three fined-grained classification datasets, i.e., CUB-200-2011 \cite{wah2011caltech}, Stanford Dogs \cite{khosla2011novel}, and FGVC-Aircraft \cite{maji2013fine}, two tabular datasets, i.e., Arcene \cite{Dua:2019}, and Iris \cite{Dua:2019}.

\textbf{Architectures} To examine whether FER works on different modern architectures, we adopt ResNet \cite{he2016deep,he2016identity}, WRN \cite{Zagoruyko2016WRN}, ShuffleNet \cite{ma2018shufflenet}, and MobileNet \cite{sandler2018mobilenetv2}.

\textbf{Baselines} As FER is modified from the the standard training procedure (STD), STD is considered as one of the most important baselines.
In addition, since FER uses soft distribution, we also compare FER with label smoothing based methods including cost-comparable self-distillation approaches, i.e., LSR \cite{muller2019does,szegedy2016rethinking}, Max-Entropy \cite{pereyra2017regularizing}, SD \cite{yang2019snapshot}, CS-KD \cite{yun2020regularizing}, and TF-Reg \cite{yuan2020revisiting}.
The comparison and discussion with LWR \cite{deng2021learning} are given in the above ablation studies.

More specific details regarding these datasets, baselines, and hyper-parameters are given in Appendix due to space limitation.

\subsubsection{Regular Classification} The regular classification is conducted on CIFAR-100 and Tiny ImageNet.
The comparison results are reported in Table \ref{regular}.
It is clearly observed that by simply restricting the DNN behavior changes on correctly-classified samples to maintain correct local boundaries, FER consistently outperforms the standard training procedure (i.e., STD) by a large margin with different network architectures across different datasets, e.g., the absolute accuracy improvements with ResNet-18 are 2.98\% and 2.15\% on CIFAR-100 and Tiny ImageNet, respectively, which demonstrates the effectiveness of FER.
On the other hand, FER also beats the other competitors significantly with different architectures on different datsets, which demonstrates the superiority of FER over these label smoothing based approaches.

\begin{table*}[!t]
\centering
\begin{tabular}{llllllll}
\toprule
Noise                 & Accuracy & STD                & LSR            & SD            & CS-KD   & TF-Reg        & FER(Ours)      \\ \midrule
\multirow{2}{*}{30\%} &Last &57.22$\pm$0.15  &60.01$\pm$0.41 &60.10$\pm$0.30 &53.86$\pm$0.12 &60.21$\pm$0.15    &\textbf{67.94$\pm$0.22 ($\uparrow$ 10.72)} \\
                      &Best &65.99$\pm$0.10 &67.38$\pm$0.33 &60.57$\pm$0.14 &54.22$\pm$0.36 &67.72$\pm$0.09   &\textbf{71.75$\pm$0.26 ($\uparrow$ 5.76)} \\ \midrule
\multirow{2}{*}{60\%} & Last  &32.22$\pm$0.32 &32.92$\pm$0.16 &32.43$\pm$0.40  &28.68$\pm$0.34 &33.19$\pm$0.40    &\textbf{61.57$\pm$0.08 ($\uparrow$ 29.35)}\\
                      & Best &53.81$\pm$0.21 &55.50$\pm$0.35 &47.70$\pm$0.12 &40.54$\pm$1.29 &54.95$\pm$0.37     &\textbf{61.99$\pm$0.21 ($\uparrow$ 8.18)}\\ \midrule
\multirow{2}{*}{80\%} & Last &12.37$\pm$0.35 &12.08$\pm$0.44 &16.49$\pm$1.27 &9.61$\pm$0.05 &12.51$\pm$0.26      &\textbf{42.74$\pm$0.77 ($\uparrow$ 30.37)} \\
                      & Best  &37.40$\pm$0.21 &36.78$\pm$1.36 &29.51$\pm$0.35 &20.31$\pm$0.53 & 38.22$\pm$0.70   &\textbf{43.36$\pm$0.63 ($\uparrow$ 5.96)} \\ \bottomrule
\end{tabular}%
\vspace{-0.05in}
\caption{Robustness results with ResNet-18 on CIFAR-100 with different levels of noise.}
\label{robust}
\end{table*}

\begin{table*}[t]
\centering
\begin{tabular}{l|cccccc}
\toprule
    Cross-dataset                                  & STD & LSR &SD & CS-KD & TF-Reg & FER (Ours) \\ \midrule
CIFAR-100 to STL-10         &69.71$\pm$0.14 &68.85$\pm$0.23 &69.27$\pm$0.13 &69.96$\pm$0.02 &69.13$\pm$0.07 &\textbf{70.80$\pm$0.15}   \\ \midrule
CIFAR-100 to Tiny ImageNet & 33.47$\pm$0.25 &32.66$\pm$0.10& 31.87$\pm$0.05 &33.46$\pm$0.05 &33.07$\pm$0.07 &\textbf{34.20$\pm$0.12}
\\ \bottomrule
\end{tabular}
\vspace{-0.05in}
\caption{Transferability performances.}
\label{transfer}
\vspace{-0.2in}
\end{table*}
\subsubsection{Fine-Grained Classification}
We further evaluate FER on fine-grained classification which is a more challenging task due to the small intra-class differences but large inter-class differences.
Table \ref{fine} reports the comparison results on three fine-grained classification benchmark datsets.
We observe that the superiority of FER over the baselines is more obvious on the fine-grained datasets than on regular datasets, e.g., the absolute accuracy improvements over STD with ResNet-18 are 6.56\%, 4.08\%, and 2.80\% on CUB, Standford Dogs, and FGVC-Aircraft, respectively.
The reason is that the samples in different classes of the fine-grained classification share more visual similarities, which leads to more wrong flipping.
FER mitigates this issue by maintaining local correct boundaries and thus obtains a much better performance.


\subsubsection{Classification on Non-Image Data}
To investigate the applicability of FER on non-image data, we conduct several experiments on two tabular datasets, i.e., Arcene and Iris.
We follow \cite{zhang2017mixup} to adopt the DNN with two hidden, fully-connected layers of 128 units.
Table \ref{tabular} reports the comparison results.
FER improves the accuracy over STD by 3.33\% and 5.56\% on Arcene and Iris, respectively, and also outperforms the other competitors substantially, which validates the applicability and usefulness of FER on non-image data.

\subsection{Robustness}
With DNNs being applied to more and more safety-critical domains such as autonomous driving and medical diagnosis, the robustness of DNNs to data noise becomes extraordinarily essential.
We follow the existing study \cite{zhang2016understanding} to evaluate the robustness of different approaches to different levels of noise.
We use the open-source implementation \footnote{https://github.com/pluskid/fitting-random-labels} of \cite{zhang2016understanding} to replace 30\%, 60\%, and 80\% of the training labels in CIFAR-100 by random noise to generate three different levels of noisy data.
Note that all the test labels are kept intact for evaluation.
When the data contain a large amount of noise, the labels are not reliable anymore and thus cannot be used to judge whether a sample has been classified correctly.
FER thus trains DNNs on noisy data by simply using Eq. (\ref{eq4}).

The robustness performances of different approaches are summarized in Table \ref{robust} where the test accuracy in the last epoch and the best test accuracy achieved during the whole training process are reported.
We have the following observations.
First, for most approaches, there is a large gap between the last accuracy and the best accuracy.
The reason is that as the training progresses, the DNN overfits a large amount of noise in the later epochs, which results in the accuracy decrease in the later epochs.
The existing studies \cite{arpit2017closer,han2018co, deng2021learning} have a similar observation that DNNs first fit training data with clean labels and then memorize those with noisy labels.
Second, FER outperforms STD and the other label smoothing baselines substantially, e.g., the last accuracy improvement over STD under 60\% noise achieves 29.35\%.
The reason is that STD totally believes in the one-hot labels and fits substantial noise in the later epochs while FER uses the weighted average behavior accumulated from the previous epochs to guide the training instead of fitting the noise, which enables it to have a much better robustness.


\subsection{Transferability}
We investigate the transferability of the representations learned by FER, as one goal of deep representation learning is to learn general representations that can be transferred to different datasets.
Specifically, we first train ResNet-18 on CIFAR-100, and then freeze the feature encoder and train a linear classifier on STL-10 \cite{coates2011analysis} and Tiny ImageNet.
All images in STL-10 and Tiny ImageNet are resized to 32$\times$32 to match the input size of the ResNet-18 pretrained on CIFAR-100.

The transferability performances are reported in Table \ref{transfer}.
The representations learned by FER have a much better cross-dataset performance than those of the other baselines.
The reason can be that FER takes into account the sample-to-class-margin distance information when learning the representations, thus avoiding overfitting the dataset and leading to more general representations.

\section{Conclusion}
In this paper, we study a novel problem about ``how many misclassified unseen samples in the last epoch were ever correctly classified by the DNN?" and then propose a simple yet effective framework FER to mitigate the wrong flipping issue, which is to our best knowledge the first work to address this problem. 
This is achieved by restricting the DNN behavior changes on correctly-classified samples with the goal of maintaining the correct local classification boundaries to avoid wrongly flipping on unseen samples.
Extensive experiments on several benchmark datasets with different modern network architectures demonstrate that FER consistently improves the generalization, the robustness, and the transferability of DNNs substantially.

\bibliography{ferm.bib}

\begin{thebibliography}{43}
\providecommand{\natexlab}[1]{#1}

\bibitem[{Arpit et~al.(2017)Arpit, Jastrzebski, Ballas, Krueger, Bengio,
  Kanwal, Maharaj, Fischer, Courville, Bengio et~al.}]{arpit2017closer}
Arpit, D.; Jastrzebski, S.~K.; Ballas, N.; Krueger, D.; Bengio, E.; Kanwal,
  M.~S.; Maharaj, T.; Fischer, A.; Courville, A.~C.; Bengio, Y.; et~al. 2017.
\newblock A Closer Look at Memorization in Deep Networks.
\newblock In \emph{ICML}.

\bibitem[{Bagherinezhad et~al.(2018)Bagherinezhad, Horton, Rastegari, and
  Farhadi}]{bagherinezhad2018label}
Bagherinezhad, H.; Horton, M.; Rastegari, M.; and Farhadi, A. 2018.
\newblock Label refinery: Improving imagenet classification through label
  progression.
\newblock \emph{arXiv preprint arXiv:1805.02641}.

\bibitem[{Chorowski and Jaitly(2016)}]{chorowski2016towards}
Chorowski, J.; and Jaitly, N. 2016.
\newblock Towards better decoding and language model integration in sequence to
  sequence models.
\newblock \emph{arXiv preprint arXiv:1612.02695}.

\bibitem[{Coates, Ng, and Lee(2011)}]{coates2011analysis}
Coates, A.; Ng, A.; and Lee, H. 2011.
\newblock An analysis of single-layer networks in unsupervised feature
  learning.
\newblock In \emph{Proceedings of the fourteenth international conference on
  artificial intelligence and statistics}, 215--223. JMLR Workshop and
  Conference Proceedings.

\bibitem[{Deng and Zhang(2021{\natexlab{a}})}]{deng2021comprehensive}
Deng, X.; and Zhang, Z. 2021{\natexlab{a}}.
\newblock Comprehensive Knowledge Distillation with Causal Intervention.
\newblock \emph{Advances in Neural Information Processing Systems}, 34.

\bibitem[{Deng and Zhang(2021{\natexlab{b}})}]{deng2021graph}
Deng, X.; and Zhang, Z. 2021{\natexlab{b}}.
\newblock Graph-Free Knowledge Distillation for Graph Neural Networks.
\newblock In \emph{The 30th International Joint Conference on Artificial
  Intelligence}.

\bibitem[{Deng and Zhang(2021{\natexlab{c}})}]{deng2021learning}
Deng, X.; and Zhang, Z. 2021{\natexlab{c}}.
\newblock Learning with Retrospection.
\newblock In \emph{Proceedings of the AAAI Conference on Artificial
  Intelligence}, volume~35, 7201--7209.

\bibitem[{Dua and Graff(2017)}]{Dua:2019}
Dua, D.; and Graff, C. 2017.
\newblock {UCI} Machine Learning Repository.

\bibitem[{Furlanello et~al.(2018)Furlanello, Lipton, Tschannen, Itti, and
  Anandkumar}]{furlanello2018born}
Furlanello, T.; Lipton, Z.~C.; Tschannen, M.; Itti, L.; and Anandkumar, A.
  2018.
\newblock Born again neural networks.
\newblock \emph{arXiv preprint arXiv:1805.04770}.

\bibitem[{Han et~al.(2018)Han, Yao, Yu, Niu, Xu, Hu, Tsang, and
  Sugiyama}]{han2018co}
Han, B.; Yao, Q.; Yu, X.; Niu, G.; Xu, M.; Hu, W.; Tsang, I.; and Sugiyama, M.
  2018.
\newblock Co-teaching: Robust training of deep neural networks with extremely
  noisy labels.
\newblock In \emph{Advances in neural information processing systems},
  8527--8537.

\bibitem[{He et~al.(2016{\natexlab{a}})He, Zhang, Ren, and Sun}]{he2016deep}
He, K.; Zhang, X.; Ren, S.; and Sun, J. 2016{\natexlab{a}}.
\newblock Deep residual learning for image recognition.
\newblock In \emph{Proceedings of the IEEE conference on computer vision and
  pattern recognition}, 770--778.

\bibitem[{He et~al.(2016{\natexlab{b}})He, Zhang, Ren, and
  Sun}]{he2016identity}
He, K.; Zhang, X.; Ren, S.; and Sun, J. 2016{\natexlab{b}}.
\newblock Identity mappings in deep residual networks.
\newblock In \emph{European conference on computer vision}, 630--645. Springer.

\bibitem[{Hinton, Vinyals, and Dean(2015)}]{hinton2015distilling}
Hinton, G.; Vinyals, O.; and Dean, J. 2015.
\newblock Distilling the knowledge in a neural network.
\newblock \emph{arXiv preprint arXiv:1503.02531}.

\bibitem[{Huang et~al.(2017)Huang, Li, Pleiss, Liu, Hopcroft, and
  Weinberger}]{huang2017snapshot}
Huang, G.; Li, Y.; Pleiss, G.; Liu, Z.; Hopcroft, J.~E.; and Weinberger, K.~Q.
  2017.
\newblock Snapshot ensembles: Train 1, get m for free.
\newblock \emph{arXiv preprint arXiv:1704.00109}.

\bibitem[{Huang et~al.(2019)Huang, Cheng, Bapna, Firat, Chen, Chen, Lee, Ngiam,
  Le, Wu et~al.}]{huang2019gpipe}
Huang, Y.; Cheng, Y.; Bapna, A.; Firat, O.; Chen, D.; Chen, M.; Lee, H.; Ngiam,
  J.; Le, Q.~V.; Wu, Y.; et~al. 2019.
\newblock Gpipe: Efficient training of giant neural networks using pipeline
  parallelism.
\newblock In \emph{Advances in neural information processing systems},
  103--112.

\bibitem[{Ji et~al.(2021)Ji, Shin, Hwang, Park, and Moon}]{ji2021refine}
Ji, M.; Shin, S.; Hwang, S.; Park, G.; and Moon, I.-C. 2021.
\newblock Refine Myself by Teaching Myself: Feature Refinement via
  Self-Knowledge Distillation.
\newblock In \emph{Proceedings of the IEEE/CVF Conference on Computer Vision
  and Pattern Recognition}, 10664--10673.

\bibitem[{Khosla et~al.(2011)Khosla, Jayadevaprakash, Yao, and
  Li}]{khosla2011novel}
Khosla, A.; Jayadevaprakash, N.; Yao, B.; and Li, F.-F. 2011.
\newblock Novel dataset for fine-grained image categorization: Stanford dogs.
\newblock In \emph{Proc. CVPR Workshop on Fine-Grained Visual Categorization
  (FGVC)}.

\bibitem[{Krizhevsky and Hinton(2009)}]{krizhevsky2009learning}
Krizhevsky, A.; and Hinton, G. 2009.
\newblock Learning multiple layers of features from tiny images.
\newblock Technical report, Citeseer.

\bibitem[{Levinson et~al.(2011)Levinson, Askeland, Becker, Dolson, Held,
  Kammel, Kolter, Langer, Pink, Pratt et~al.}]{levinson2011towards}
Levinson, J.; Askeland, J.; Becker, J.; Dolson, J.; Held, D.; Kammel, S.;
  Kolter, J.~Z.; Langer, D.; Pink, O.; Pratt, V.; et~al. 2011.
\newblock Towards fully autonomous driving: Systems and algorithms.
\newblock In \emph{2011 IEEE Intelligent Vehicles Symposium (IV)}, 163--168.
  IEEE.

\bibitem[{Loshchilov and Hutter(2017)}]{loshchilov2016sgdr}
Loshchilov, I.; and Hutter, F. 2017.
\newblock Sgdr: Stochastic gradient descent with warm restarts.
\newblock \emph{International Conference for Learning Representations}.

\bibitem[{Ma et~al.(2018)Ma, Zhang, Zheng, and Sun}]{ma2018shufflenet}
Ma, N.; Zhang, X.; Zheng, H.-T.; and Sun, J. 2018.
\newblock Shufflenet v2: Practical guidelines for efficient cnn architecture
  design.
\newblock In \emph{Proceedings of the European conference on computer vision
  (ECCV)}, 116--131.

\bibitem[{Maji et~al.(2013)Maji, Rahtu, Kannala, Blaschko, and
  Vedaldi}]{maji2013fine}
Maji, S.; Rahtu, E.; Kannala, J.; Blaschko, M.; and Vedaldi, A. 2013.
\newblock Fine-grained visual classification of aircraft.
\newblock \emph{arXiv preprint arXiv:1306.5151}.

\bibitem[{Miotto et~al.(2016)Miotto, Li, Kidd, and Dudley}]{miotto2016deep}
Miotto, R.; Li, L.; Kidd, B.~A.; and Dudley, J.~T. 2016.
\newblock Deep patient: an unsupervised representation to predict the future of
  patients from the electronic health records.
\newblock \emph{Scientific reports}, 6(1): 1--10.

\bibitem[{M{\"u}ller, Kornblith, and Hinton(2019)}]{muller2019does}
M{\"u}ller, R.; Kornblith, S.; and Hinton, G.~E. 2019.
\newblock When does label smoothing help?
\newblock In \emph{Advances in Neural Information Processing Systems},
  4694--4703.

\bibitem[{Ozbayoglu, Gudelek, and Sezer(2020)}]{ozbayoglu2020deep}
Ozbayoglu, A.~M.; Gudelek, M.~U.; and Sezer, O.~B. 2020.
\newblock Deep learning for financial applications: A survey.
\newblock \emph{Applied Soft Computing}, 93: 106384.

\bibitem[{Pereyra et~al.(2017)Pereyra, Tucker, Chorowski, Kaiser, and
  Hinton}]{pereyra2017regularizing}
Pereyra, G.; Tucker, G.; Chorowski, J.; Kaiser, {\L}.; and Hinton, G. 2017.
\newblock Regularizing neural networks by penalizing confident output
  distributions.
\newblock \emph{arXiv preprint arXiv:1701.06548}.

\bibitem[{Real et~al.(2019)Real, Aggarwal, Huang, and Le}]{real2019regularized}
Real, E.; Aggarwal, A.; Huang, Y.; and Le, Q.~V. 2019.
\newblock Regularized evolution for image classifier architecture search.
\newblock In \emph{Proceedings of the aaai conference on artificial
  intelligence}, volume~33, 4780--4789.

\bibitem[{Sandler et~al.(2018)Sandler, Howard, Zhu, Zhmoginov, and
  Chen}]{sandler2018mobilenetv2}
Sandler, M.; Howard, A.; Zhu, M.; Zhmoginov, A.; and Chen, L.-C. 2018.
\newblock Mobilenetv2: Inverted residuals and linear bottlenecks.
\newblock In \emph{Proceedings of the IEEE conference on computer vision and
  pattern recognition}, 4510--4520.

\bibitem[{Simonyan and Zisserman(2015)}]{simonyan2014very}
Simonyan, K.; and Zisserman, A. 2015.
\newblock Very deep convolutional networks for large-scale image recognition.
\newblock In \emph{International Conference for Learning Representations}.

\bibitem[{Szegedy et~al.(2016)Szegedy, Vanhoucke, Ioffe, Shlens, and
  Wojna}]{szegedy2016rethinking}
Szegedy, C.; Vanhoucke, V.; Ioffe, S.; Shlens, J.; and Wojna, Z. 2016.
\newblock Rethinking the inception architecture for computer vision.
\newblock In \emph{Proceedings of the IEEE conference on computer vision and
  pattern recognition}, 2818--2826.

\bibitem[{Toneva et~al.(2019)Toneva, Sordoni, Combes, Trischler, Bengio, and
  Gordon}]{toneva2018empirical}
Toneva, M.; Sordoni, A.; Combes, R. T.~d.; Trischler, A.; Bengio, Y.; and
  Gordon, G.~J. 2019.
\newblock An empirical study of example forgetting during deep neural network
  learning.
\newblock In \emph{International Conference for Learning Representations}.

\bibitem[{Vaswani et~al.(2017)Vaswani, Shazeer, Parmar, Uszkoreit, Jones,
  Gomez, Kaiser, and Polosukhin}]{vaswani2017attention}
Vaswani, A.; Shazeer, N.; Parmar, N.; Uszkoreit, J.; Jones, L.; Gomez, A.~N.;
  Kaiser, {\L}.; and Polosukhin, I. 2017.
\newblock Attention is all you need.
\newblock In \emph{Advances in neural information processing systems},
  5998--6008.

\bibitem[{Wah et~al.(2011)Wah, Branson, Welinder, Perona, and
  Belongie}]{wah2011caltech}
Wah, C.; Branson, S.; Welinder, P.; Perona, P.; and Belongie, S. 2011.
\newblock The caltech-ucsd birds-200-2011 dataset.
\newblock \emph{California Institute of Technology}.

\bibitem[{Xu and Liu(2019)}]{xu2019data}
Xu, T.-B.; and Liu, C.-L. 2019.
\newblock Data-distortion guided self-distillation for deep neural networks.
\newblock In \emph{Proceedings of the AAAI Conference on Artificial
  Intelligence}, volume~33, 5565--5572.

\bibitem[{Yang et~al.(2019{\natexlab{a}})Yang, Xie, Qiao, and
  Yuille}]{yang2019training}
Yang, C.; Xie, L.; Qiao, S.; and Yuille, A.~L. 2019{\natexlab{a}}.
\newblock Training deep neural networks in generations: A more tolerant teacher
  educates better students.
\newblock In \emph{Proceedings of the AAAI Conference on Artificial
  Intelligence}, volume~33, 5628--5635.

\bibitem[{Yang et~al.(2019{\natexlab{b}})Yang, Xie, Su, and
  Yuille}]{yang2019snapshot}
Yang, C.; Xie, L.; Su, C.; and Yuille, A.~L. 2019{\natexlab{b}}.
\newblock Snapshot distillation: Teacher-student optimization in one
  generation.
\newblock In \emph{Proceedings of the IEEE Conference on Computer Vision and
  Pattern Recognition}, 2859--2868.

\bibitem[{Yuan et~al.(2020)Yuan, Tay, Li, Wang, and Feng}]{yuan2020revisiting}
Yuan, L.; Tay, F.~E.; Li, G.; Wang, T.; and Feng, J. 2020.
\newblock Revisiting Knowledge Distillation via Label Smoothing Regularization.
\newblock In \emph{Proceedings of the IEEE/CVF Conference on Computer Vision
  and Pattern Recognition}, 3903--3911.

\bibitem[{Yun et~al.(2020)Yun, Park, Lee, and Shin}]{yun2020regularizing}
Yun, S.; Park, J.; Lee, K.; and Shin, J. 2020.
\newblock Regularizing class-wise predictions via self-knowledge distillation.
\newblock In \emph{Proceedings of the IEEE/CVF Conference on Computer Vision
  and Pattern Recognition}, 13876--13885.

\bibitem[{Zagoruyko and Komodakis(2016)}]{Zagoruyko2016WRN}
Zagoruyko, S.; and Komodakis, N. 2016.
\newblock Wide Residual Networks.
\newblock In \emph{BMVC}.

\bibitem[{Zhang et~al.(2017)Zhang, Bengio, Hardt, Recht, and
  Vinyals}]{zhang2016understanding}
Zhang, C.; Bengio, S.; Hardt, M.; Recht, B.; and Vinyals, O. 2017.
\newblock Understanding deep learning requires rethinking generalization.
\newblock \emph{International Conference for Learning Representations}.

\bibitem[{Zhang et~al.(2018)Zhang, Cisse, Dauphin, and
  Lopez-Paz}]{zhang2017mixup}
Zhang, H.; Cisse, M.; Dauphin, Y.~N.; and Lopez-Paz, D. 2018.
\newblock mixup: Beyond empirical risk minimization.
\newblock \emph{International Conference for Learning Representations}.

\bibitem[{Zhang et~al.(2019)Zhang, Song, Gao, Chen, Bao, and
  Ma}]{zhang2019your}
Zhang, L.; Song, J.; Gao, A.; Chen, J.; Bao, C.; and Ma, K. 2019.
\newblock Be your own teacher: Improve the performance of convolutional neural
  networks via self distillation.
\newblock In \emph{Proceedings of the IEEE International Conference on Computer
  Vision}, 3713--3722.

\bibitem[{Zoph et~al.(2018)Zoph, Vasudevan, Shlens, and Le}]{zoph2018learning}
Zoph, B.; Vasudevan, V.; Shlens, J.; and Le, Q.~V. 2018.
\newblock Learning transferable architectures for scalable image recognition.
\newblock In \emph{Proceedings of the IEEE conference on computer vision and
  pattern recognition}, 8697--8710.

\end{thebibliography}

\end{document}